%% file: main.tex
\documentclass[10pt,conference]{IEEEtran}
\usepackage{amsmath,amssymb,amsfonts}
\usepackage{soul}
\usepackage{algorithmic}
\usepackage{graphicx}
\usepackage[lofdepth,lotdepth]{subfig}
\graphicspath{ {figures/} }
\usepackage{xcolor}
\usepackage{todonotes}
\usepackage{cite}
\usepackage{url}
\usepackage{multicol}
\usepackage[letterpaper,top=25.4mm,bottom=28.6mm,left=20mm,right=20mm]{geometry}
\begin{document}


\title{\vspace{10mm}\fontsize{16pt}{20pt}\selectfont \textbf{Large-Scale Object Detection of Images from Network Cameras in Variable Ambient Lighting Conditions}}

\author{\IEEEauthorblockN{Caleb Tung, Matthew R. Kelleher, Ryan J. Schlueter, Binhan Xu,\\
Yung-Hsiang Lu, George K. Thiruvathukal\IEEEauthorrefmark{1}\IEEEauthorrefmark{2}, Yen-Kuang Chen\IEEEauthorrefmark{3},
Yang Lu\IEEEauthorrefmark{4}
}
\IEEEauthorblockA{Purdue University, School of Electrical and Computer Engineering, 
West Lafayette, IN, USA\\
\IEEEauthorrefmark{1}Loyola University Chicago, Department of Computer Science, Chicago, IL, USA\\
\IEEEauthorrefmark{2}Argonne National Laboratory, Argonne Leadership Computing Facility, Argonne, IL, USA\\
\IEEEauthorrefmark{3}Intel Corporation, Santa Clara, CA, USA}
\IEEEauthorrefmark{4}Facebook Corporation, Menlo Park, CA, USA}
\maketitle

\input{abstract}

\vspace*{-\baselineskip}
\vspace{12mm}
\input{introduction}

\input{method}

\input{results}

\section{Opportunities for Improvement}
\input{discussion}

\section{Conclusion}

\input{conclusion}

\section{Acknowledgments}
\input{acknowledgments}


\bibliographystyle{unsrt}
\bibliography{1bie}


\end{document}

%% file: abstract.tex
\begin{abstract}


Computer vision relies on labeled datasets for training and evaluation in detecting and recognizing objects. The popular computer vision program, YOLO (``You Only Look Once''), has been shown to accurately detect objects in many major image datasets. However, the images found in those datasets, are independent of one another and cannot be used to test YOLO's consistency at detecting the same object as its environment (e.g. ambient lighting) changes. This paper describes a novel effort to evaluate YOLO's consistency for large-scale applications. It does so by working (a) at large scale and (b) by using consecutive images from a curated network of public video cameras deployed in a variety of real-world situations, including traffic intersections, national parks, shopping malls, university campuses, etc. We specifically examine YOLO's ability to detect objects in different scenarios (e.g., daytime vs. night), leveraging the cameras' ability to rapidly retrieve many successive images for evaluating detection consistency. Using our camera network and advanced computing resources (supercomputers), we analyzed more than 5 million images captured by 140 network cameras in 24 hours. Compared with labels marked by humans (considered as ``ground truth''), YOLO struggles to consistently detect the same humans and cars as their positions change from one frame to the next; it also struggles to detect objects at night time. Our findings suggest that state-of-the art vision solutions should be trained by data from network camera with contextual information before they can be deployed in applications that demand high consistency on object detection.

\end{abstract}

%% file: introduction.tex
\section{Introduction and Related Work}



Computer vision technologies  have been widely used in many systems such as airport security
and autonomous vehicles.  Vehicle collision avoidance analyzes visual data to detect the presence of pedestrians, other cars, or lane dividing lines~\cite{Eyesight_page}. To protect privacy in photos, a phone app may detect human faces and blur the regions~\cite{Snapchat_patent}.  These are examples of {\it object detection}. Learning-based vision technologies relies on image datasets for training and evaluation~\cite{LeCun2015NatureDeep}. The characteristics of the images play crucial roles in the success of vision technologies~\cite{IRI_paper}. Commonly used datasets such as ImageNet~\cite{ImageNet_paper}, COCO (Common Object in Context)~\cite{COCO_paper}, PASCAL VOC (Pattern Analysis, Statistical Modeling, and Computational Learning Visual Object Classes)~\cite{PascalVOC_paper}, and INRIA~\cite{INRIA_paper} do not have ``context" (such as time and location). In contrast, AMOS~\cite{AMOS_paper} retrieves data from network cameras and can observe seasonal changes. AMOS uses only low refresh rates.

These datasets are sufficient for confirming learning-based object detection for certain applications. However, when used for large-scale, 24-hour video stream recognition, the datasets may become unreliable metrics of an object detector's performance. Running all day and at large-scale, applications demand high consistency. Consider a hypothetical piece of software that runs round-the-clock image detection on traffic intersection for emergencies. This application demands accurate, and as importantly, \textit{consistent} object detection, regardless of changing lighting conditions.
Thus, consistently detecting an object from one frame to the next becomes critical. This requirement currently cannot be verified with the independent images in existing datasets. Changing lighting conditions and moving shadows would add to the complexity of the problem. Even indoor scenes could change dramatically, as people turn lights on and off.

Therefore, before any large-scale, fulltime object detection application is adopted, object detectors must be carefully evaluated to see how well they can generalize to the changes introduced into the images. This paper presents a study evaluating and validating a state-of-the-art vision
solution that could be expected from outdoor/indoor image streams, to identify modifications that may need to be made before the detectors can be deployed in such a manner.

\begin{figure}[h]
  \centering
  \includegraphics[width=\linewidth]{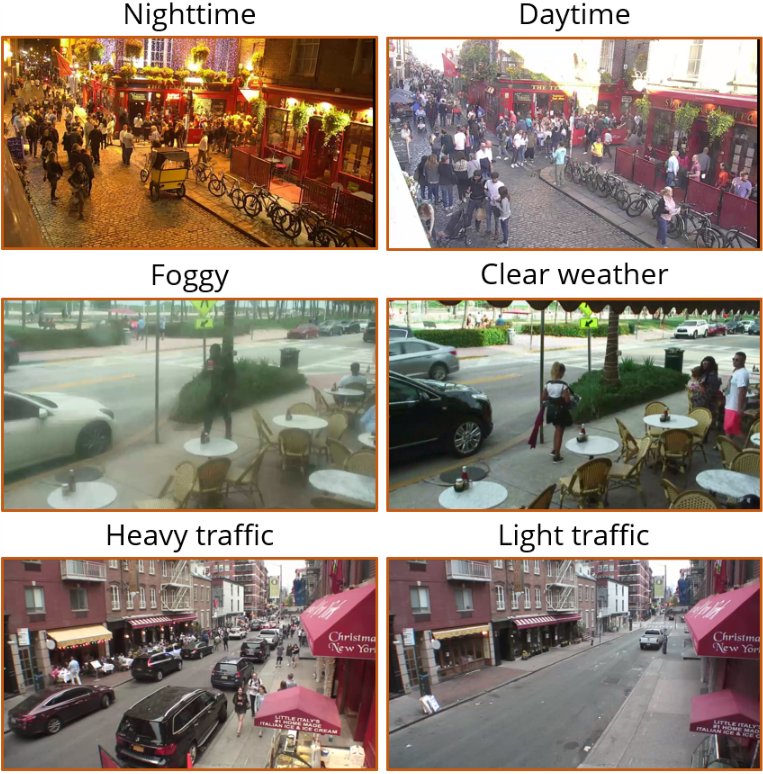}
  \vspace{-0.1in}
  \caption{Network cameras can capture images taken by the same cameras as different time of day. Note that fog is not addressed by this paper, as it is difficult to quantify.}
  \label{fig:different-scenarios}
\end{figure}

To the authors' knowledge, this is the largest study (5 million of images) evaluating whether object detection can be performed consistently on images taken by the same cameras in different contextual scenarios (such as ambient lighting). 
This study uses network cameras (instead of handheld cameras) to ensure that the background is consistent.
Figure~\ref{fig:different-scenarios} shows three pairs of images taken by the same cameras at different time.



YOLO (You Only Look Once) is considered one of the state-of-the-art computer vision solutions~\cite{TrafficLight_paper}. Redmon, \textit{et al.}~\cite{YOLO_site} found that YOLO outperformed several object detection methods, such as Single-Shot Detector, in terms of both accuracy and speed~\cite{SSD_paper,YOLO_paper}.  The comparison uses the  COCO and PASCAL VOC datasets~\cite{YOLO_paper}. Neither dataset includes images from the same camera for the purpose of evaluating consistency and YOLO's ability to handle images taken at night.
YOLO is used in many computer vision applications. For example, Jensen \textit{et al.}  used YOLO as a stoplight detector for autonomous vehicles~\cite{TrafficLight_paper}. Al-Masni \textit{et al.} designed a YOLO-based malignant tumor learning model for breast cancer detection~\cite{BreastCancer_paper}. Xie \textit{et al.} developed a license plate detector~\cite{LicensePlate_paper}. Kang et al.~\cite{8342102} improved YOLO for the winning solution in the 2017 Low-Power Image Recognition Challenge. In all of these cases, however, the images did not include any variation(s) introduced by network cameras.


This paper uses YOLO for detecting objects in images acquired from network cameras  at high refresh rate using
a research tool for computer vision~\cite{CAM2_paper}. This study downloads more than 5 million images from 140 network cameras over 24 hours (about every 2 seconds from
each camera). A small subset of these images is validated by humans. The subset is used to investigate YOLO's accuracy as the ambient lighting changes. Additionally, this study  downloads images  from the same cameras once per second to evaluate the consistency when many objects in the scenes are unchanged. 
Even though this study focuses on YOLO, the same methodology would be applicable to other vision solutions.

The contributions of this paper include: 
(1) This paper presents a study evaluating how changes in ambient lighting affects object detection.
Such a study is not possible using existing image datasets because these images are independent.
(2) This study reveals that current solutions for object detection overlook consistency in object detection
when analyzing images taken within short time intervals by the same cameras.

%% file: method.tex
\section{Methodology}
This section describes the experiments to evaluate YOLO. This section first explains (section II.A, B) how to acquire images from network cameras. Then, it describes the systems (section II.C) used to run the experiments and the procedure (section II.D) to sample the images for validation.

\subsection{Network Cameras}
Most existing datasets contain images that cover a wide variety of scenes, but the images are independent (e.g. there could be 10 photos of humans but the humans would be different in each photo, the scenery would be different, and the lighting wouldn't be changing). It is unclear how vision technologies can handle dependent images like those streamed from video cameras.
To answer this question, this paper uses real-time images from public cameras streaming at ~30 frames per second over the Internet. The images are sent in real-time. The advantages of using real-time network cameras instead of existing datasets (such as ImageNet and COCO) are: (1) It is possible to obtain images at high rates from the same cameras. Thus, this study can evaluate whether objects can be detected consistently in the same scene as they move and the environment around them is unchanged.
(2) Based on the cameras' locations, it is possible to obtain images at different time of day. Hence, this study can compare object detection at different ambient lighting. (3) Because of cameras' known locations, it is possible to predict the likelihood of objects. For example, knowing that a store is closed would mean that any computer vision running on in-store cameras could intelligently decrease the likelihood of any false human detections, since people are not usually in a store when it is closed.

This study  uses the CAM\textsuperscript{2} (Continuous Analysis of Many CAMeras) software infrastructure~\cite{CAM2_paper} built at Purdue University for retrieving images from network cameras. CAM\textsuperscript{2} curates a large database of live network cameras and allows queries by geographical location.


\subsection{Images from Network Cameras}

This study collected five million images from 140 network cameras around the world to evaluate YOLO's capability in  detecting people and cars.  This study focuses on people and cars because they are of interest in many vision applications (such as monitoring traffic and cashier-less retail stores). Network cameras 
can provide video streams at roughly 30 frames per second and the resolutions range between 480p and 1080p.
This study uses images  collected over 24 hours. Even though these cameras can provide video streams, this study focuses on image-based object detection for two reasons: (1) More image datasets (with labels) are available than video datasets.
(2) Many network cameras (in particular traffic cameras) provide only snapshots every several seconds to several minutes.


\subsection{Computing Clusters}

This study uses two computing clusters for running large-scale image recognition.  The first has 200 nodes and each CPU include Skylake ~\cite{Skylake_page} and Knights Landing~\cite{KnightsLanding_page} Xeon Phi cores. Each node has 220 GB of RAM and a 400 GB solid state drive. The entire cluster ties into a central 2 PB network file storage system. YOLO was rebuilt for the Caffe ~\cite{Caffe_page} neural network framework. In order to accommodate the large-scale data processing, the study uses a CPU-optimized version of Caffe to run on those cores. The second cluster has 128 nodes; each node has 12 CPU cores and one NVIDIA Tesla K80 dual-GPU card. This cluster totals 47 TB of system RAM and 3 TB of GPU RAM. YOLO ran on its original Darknet framework~\cite{YOLO_site}, optimized for the GPUs. The 5 million images require approximately 5 TB of storage. These resources make this large scale study possible; the images could be split across the two computing clusters and processed in parallel, allowing our experiment to properly simulate YOLO under large-scale application conditions. This paper studies only the accuracy of object detection, so comparing the execution time of the two clusters is beyond the scope of this paper.  


\subsection{Validation against Ground Truth}


This study uses new data from network cameras, and new labels must be created for validation.  Since computer vision is still a research
topic, the labels are created by humans. Labeling all 5 million images would require significant effort.  Suppose labeling one image takes 10 seconds, labeling 5 million images needs nearly 14,000 hours or 580 days.
Thus, this study selects only a small subset of images to label. The following method is adopted for selecting the images:




\begin{enumerate}
\item Four images were uniformly sampled from each of the 140 cameras at hourly intervals, over the 24-hour period. Thus, we would have images from all cameras spanning a full day. In total, 13,440 ($4 \times 140 \times 24$) were selected.

\item The 13,440 images were reviewed to eliminate a few low quality images (e.g., excessively grainy images, etc.) Only images that contain clearly visible humans and cars are selected. This was done completely by hand and resulted in the removal of 400 images.


\item The labels created by YOLO are compared with the ground-truth labels to establish YOLO's consistency and accuracy with the network camera images.
\end{enumerate}

YOLO marks each recognized object using a {\it bounding box}. The box is then compared with a human-labeled bounding box. {\it Intersection over Union} (IOU) is used to determine whether the two bounding boxes have sufficient overlap.
Suppose $B_1$ and $B_2$ are the two bounding box.  IOU is defined as

\begin{equation*}
  \text{IOU} = \frac{B_1 \cap B_2}{B_1 \cup B_2}. 
\end{equation*}

IOU of 1.0 means the two bounding boxes match perfectly. IOU above 0.5 is commonly used to indicate that objects have been successfully identified~\cite{LPIRC_paper}.






%% file: results.tex
\section{Detecting Objects in Images}

This study evaluates YOLO's accuracy detecting objects in images acquired from network cameras. The results are divided into two categories: (1) daytime vs. nighttime, and (2) consistency of adjacent frames. 

\subsection{Daytime vs. Nighttime}

\begin{figure}[!ht]
  \centering
  \includegraphics[width=\linewidth]{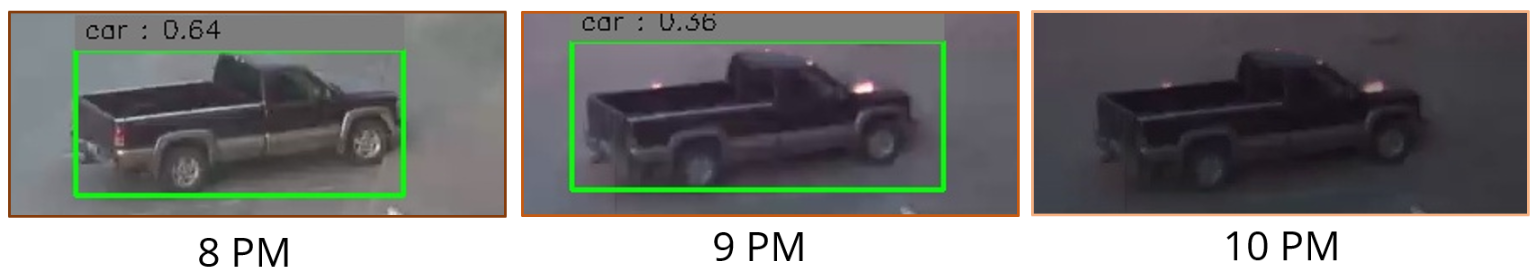}
  \caption{YOLO's IOU
  drops from 0.67 at 8PM to 0.36 at 9PM.  At 10PM, YOLO does not detect the truck.}\label{fig:detect_truck_night}
\end{figure}

Most datasets have images with abundant ambient lighting, and few vision solutions compare object detection in daytime or nighttime. One advantage of network cameras is that they can provide data 24 hours. 
This subsection describes YOLO's accuracy (IOU and confidence) detecting objects in outdoor scenes throughout the day as the amount of sunlight changes.
YOLO tends to miss objects when it gets darker.  Figure~\ref{fig:detect_truck_night} shows three images from the same camera, taken an hour apart. The black pickup truck does not move. As it gets darker, YOLO's ability to detect this truck declines. The confidence level starts at 0.8 at 8PM, dropping to 0.64 at 9PM. The IOU drops from 0.67 to 0.36. At 10PM, YOLO is unable to detect the truck when only streetlight illumination is available.

\begin{figure}[!ht]
  \centering
  \includegraphics[width=\linewidth]{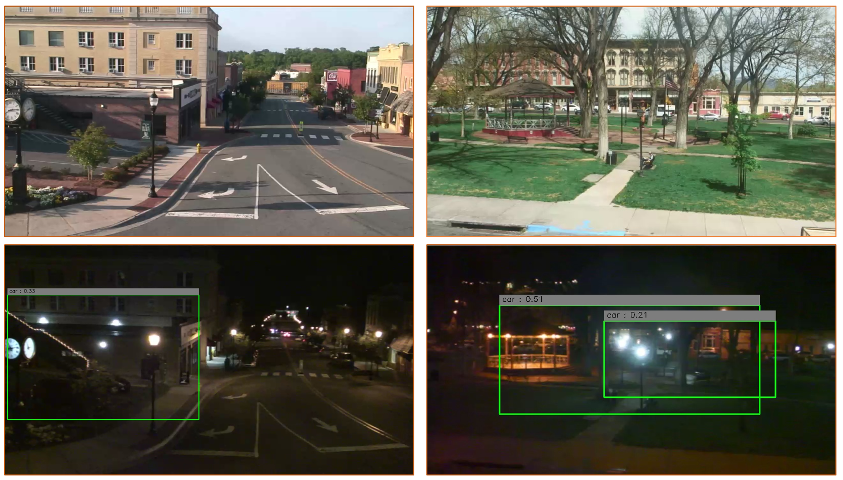}
  \caption{During the day, YOLO correctly reports no cars. At night, YOLO treats street lights as cars' headlights.}\label{fig:night-false}
\end{figure}

YOLO has  false positives in some images from other network cameras.  In Figure~\ref{fig:night-false}, 
YOLO correctly reports zero presence of people and vehicles during the day. At night, however, YOLO mistakenly
recognizes the streetlights as vehicles' headlights and reports the presence of cars.

For systematic evaluation, the authors select cameras that are likely to see many people and vehicles. One image is taken from each camera every hour. People and vehicles are marked by hand as ground truth and YOLO's results are compared with the labels. 
The classification is defined as
(1)  True Positive (TP): there is an object and YOLO detects it.
(2)  True Negative (TN): there is no object and YOLO detects nothing. 
(3) False Positive (FP): there is no object and YOLO says there is. 
(4) False Negative  (FN): there is an object and YOLO fails to detect it.
The results are shows in Table~\ref{table:night}.
\begin{table}[h]
\centering
\begin{tabular}{|l|rrrr|rrrr|} \hline
& \multicolumn{4}{|c|}{\bf Cars} & \multicolumn{4}{|c|}{\bf Humans} \\
& {\bf TP} & {\bf TN} & {\bf FP} & {\bf FN} & {\bf TP} & {\bf TN} & {\bf FP} & {\bf FN}\\ \hline
Daytime & 0.80 & 0.99 & 0.01 & 0.20 & 0.60 & 1.0 & 0 & 0.40 \\
Night & 0.12 & 0.71 & 0.29 & 0.88 & 0.06 & 1.0 & 0 & 0.94 \\ \hline
\end{tabular}
\caption{YOLO's ability to detect objects. Correct detection must have IOU $\ge$ 0.5. Only objects detected at
confidence level 0.3 or higher are counted.}
\label{table:night}
\end{table}

Note that the ``TN'' and ``FP'' values for human detection are from a limited sample and mainly indicate that YOLO is able to correctly discern the absence of objects at daytime (few false positives).
At night, however, YOLO mistakenly detects many cars that do not exist (more false positives). 










\subsection{Consistency in Object Detection }

\begin{figure}[!ht]
  \centering
   \includegraphics[width=\linewidth]{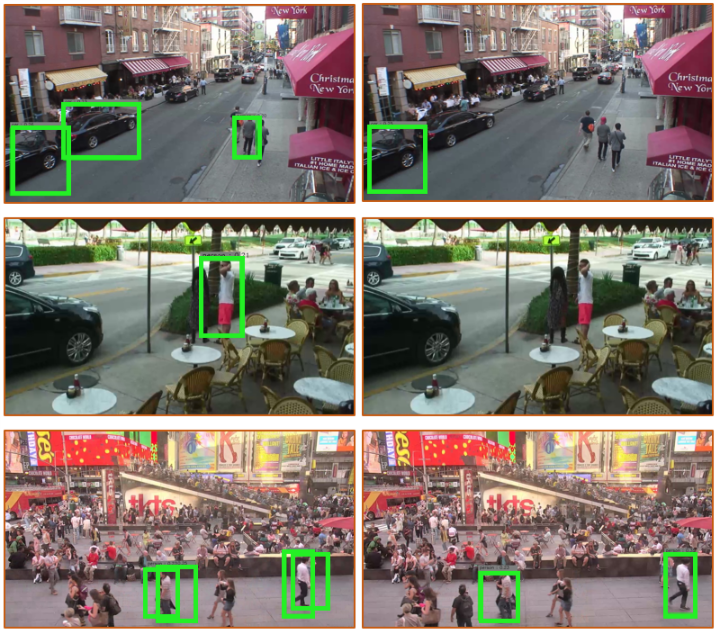}
 \caption{Example of detected objects (enclosed by bounding boxes) between adjacent frames. Top: shop-lined city street. Middle: roadside cafe. Bottom: urban sidewalk.}
 \label{fig:all-consistency}
\end{figure}

\begin{figure}[!ht]
  \centering
  \includegraphics[width=\linewidth]{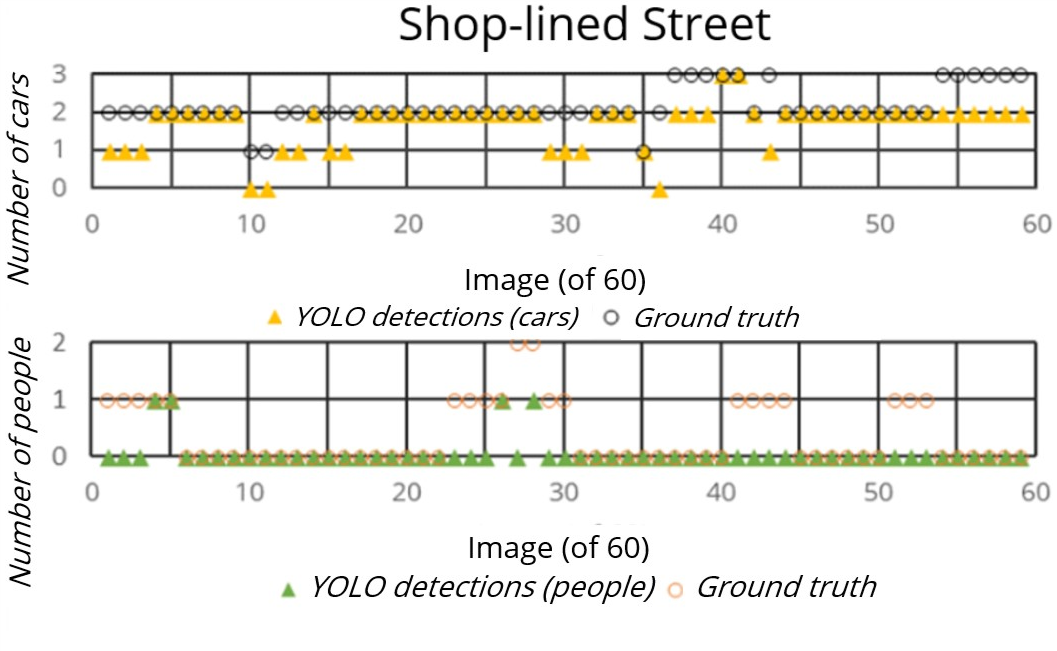}
  \includegraphics[width=\linewidth]{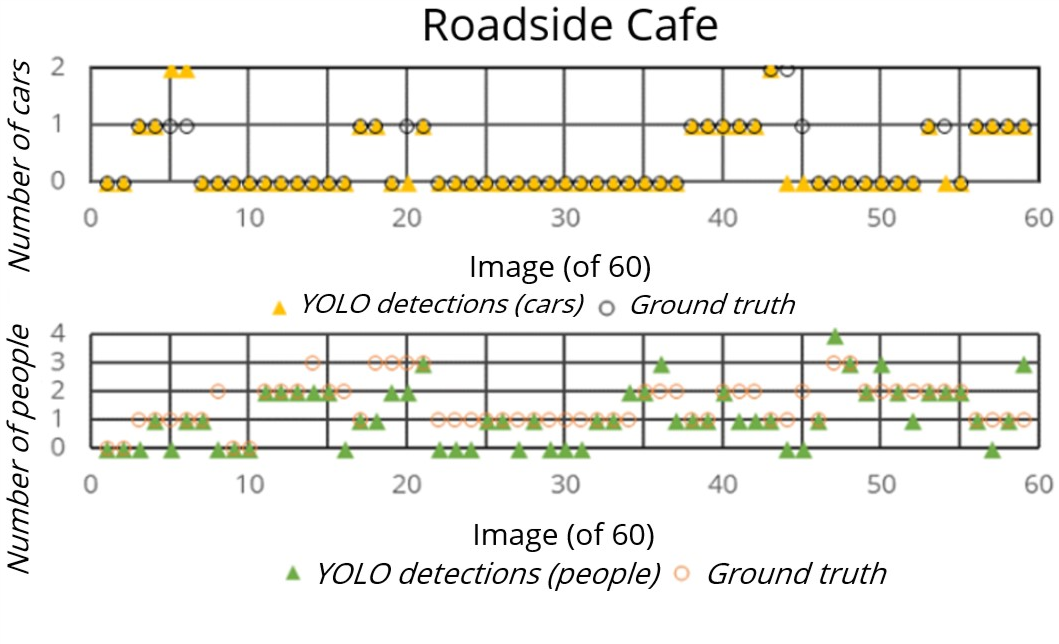}
   \includegraphics[width=\linewidth]{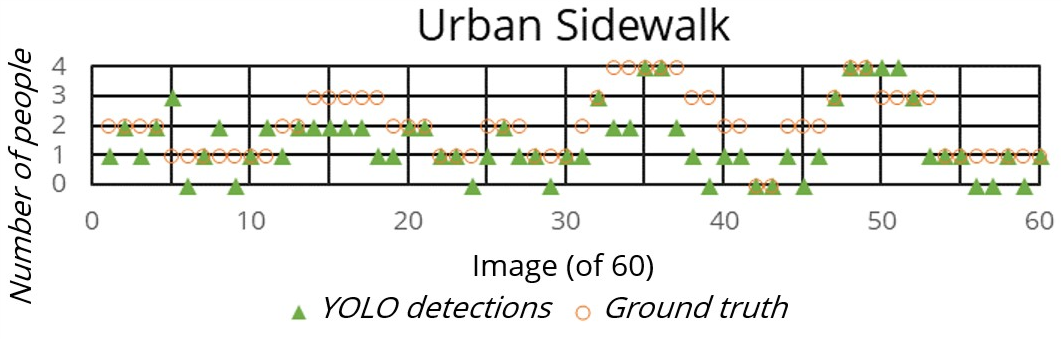}
   \setlength{\belowcaptionskip}{-10pt}
  \caption{
  The number of people and cars manually marked on each image, compared with YOLO's detections.}\label{fig:chart-miami-cafe}
\end{figure}

This paper also investigates whether YOLO consistently detects (or misses) objects in similar images taken within short time intervals. This is again something that has yet to be tested, but is very critical to any sort of large-scale application that cannot afford any lapse in consistency. The experiment uses 180 images taken every second from 3 network cameras and compares YOLO's results with ground truth.
Figure~\ref{fig:all-consistency} shows two images from these three cameras.  If an object is detected by YOLO, the bounding box is shown. From this figure, it is clear that object detection is unstable---some objects are detected in one frame and missed in the next frame. 
Figure~\ref{fig:chart-miami-cafe} compares the numbers of objects detected by YOLO with that by humans.  This figure shows
that in many cases YOLO's results are unstable. For example, in the New York alley, the number of cars stayed at 2 during the 13th and 34th seconds but YOLO's reported only one car occasionally. As another example, at the Streetside Cafe, during the 22nd and the 34th seconds, the number of people remained one. YOLO detected 0, 1, and 2 in this duration. 
Overall, YOLO is unstable at detecting objects in images one second apart on the same camera. It is worth noting that YOLO's accuracy also strayed from the numbers reported in section III.A for images taken an hour apart, further emphasizing YOLO's instability.

%% file: discussion.tex
This paper uses YOLO as the program for computer vision but the findings are not restricted to YOLO. Instead, this paper identifies opportunities for improvements in the general  methodologies for research in computer vision.

Learning-based computer vision relies on datasets for training and evaluation. Popular datasets are forced to treat each image independently because
the images are scattered across sources, not from the same cameras capturing scenes at the
same angles at different time. Thus, without using network camera images, it is practically impossible to perform the study presented in this paper. However, this study is critical to the success of a large-scale, real-time application that must give consistent detections repeatedly, no matter the time of the day and the lighting conditions.

Our prior study~\cite{IRI_paper} discovered much noticeable \emph{bias} in commonly used datasets. For example, there are few objects of interest in each image. The objects are often located at or near the centers of images and the objects tend to have high pixel resolution. In contrast, images acquired from network cameras often contain many objects of interest. These objects may appear anywhere in the images, and move around over time. This paper extends the prior work by investigating the ability of a state-of-the-art vision program (YOLO) to detect objects and suggests that new datasets are needed to consistently handle changing ambient lighting in similar images taken within short time durations.

This paper does not consider how ``context'' information (e.g. time and location) could be used in vision programs. It is highly improbable for a camera deployed in a city center to see a buffalo. Thus, if a vision program detects a buffalo, the program is likely wrong without human checking. Future work will explore how to integrate context information to assist humans in labeling, which is in turn important for large-scale image detection efforts.

%% file: conclusion.tex
This paper evaluates YOLO's performance with network camera images (1) short time intervals apart and (2) distributed equally between daytime and nighttime. For any large-scale application that runs object detection outdoors and indoors at real time, consistency and the ability to generalize even when ambient lighting changes is critically important. Using a global network of real-time cameras, this paper is able to more accurately quantitatively test the state-of-the-art object detector than other modern image datasets, granting unique insights into YOLO's performance at large-scale.
Similar to a large-scale, all-day image detection application, the paper considers several cases where the changing environment on a given camera may cause dramatic changes in YOLO's accuracy. Our analysis discovers that YOLO struggles to identify dark-colored objects and gives false positives at night. In addition, YOLO cannot consistently detect objects from the same camera. Based on the new evidence obtained from our richer image experiments, we believe further study of network camera-centric object detection is key; it will improve computer vision programs so they can support large-scale applications.

We suggest that a good solution can be found in training with a wide collection of network camera images like those we curate in the CAM\textsuperscript{2} system. These cameras provide continuous streams of real-time images that are different even at the same time of day or under identical lighting conditions. Even more importantly, network cameras also afford the luxury of context-awareness. For example, a computer vision program could rule out the presence of a car, given camera placement (e.g. facing a lake). Given the serious potential for false positives, developing a comprehensive strategy for growing a robust, context-aware training dataset is necessary to train and evaluate the next generation of computer vision solutions for large-scale, real-time applications.

%% file: acknowledgments.tex
This project is supported by NSF OAC-1535108 and OAC-1747694 and by Facebook. It uses computer resources of the Intel Labs Academic Research Program and the Argonne Leadership Computing Facility, a DOE Office of Science User Facility supported under Contract DE-AC02-06CH11357. Any opinions, findings, and conclusions or recommendations expressed in this paper are those of the authors and do not necessarily reflect the views of the sponsors. The authors want to thank Jaeyoun Kim at Google for his valuable suggestions.